\documentclass[11pt]{article} 

\usepackage[utf8]{inputenc} 

\usepackage{geometry} 
\usepackage{graphicx} 

\usepackage{booktabs} 
\usepackage{array} 
\usepackage{paralist} 
\usepackage{verbatim} 
\usepackage{subfig} 
\usepackage{amsmath}
\usepackage{amssymb}
\usepackage{amsthm}
\usepackage{setspace}
\usepackage{natbib}

\usepackage{fancyhdr} 
\usepackage{sectsty}
\allsectionsfont{\sffamily\mdseries\upshape} 

\usepackage[nottoc,notlof,notlot]{tocbibind} 
\usepackage[titles,subfigure]{tocloft} 

\newcommand{\bs}[1]{\boldsymbol#1}
\newcommand{\pd}[2]{\frac{\partial #1}{\partial #2}}
\newcommand{\expit}[1] {{(1+\exp(-\gamma #1))}^{-1}}
\newcommand{\ap}{\mathcal{A}^p(\theta,  (\bs{x}, y) ) }
\newcommand{\apii}{\mathcal{A}^p(\theta,  (\bs{x}_i, y_i) ) }
\newcommand{\api}{\mathcal{A}^p(\theta,  (\bs{x}_i, y) ) }
\newcommand{\ai}{\alpha_i(\theta)}
\newcommand{\aj}{\alpha_i(\theta)}

\theoremstyle{definition}
\newtheorem{example}{Example}

\title{Optimized conformal classification using gradient descent approximation}
\author{Anthony Bellotti\\
School of Computer Science, University of Nottingham Ningbo China\\
Correspondence: \texttt{Anthony-Graham.Bellotti@nottingham.edu.cn} }
\date{}

\begin{document}
\doublespacing
\maketitle

\begin{abstract}
Conformal predictors are an important class of algorithms that allow predictions to be made with a user-defined confidence level.
They are able to do this by outputting prediction sets, rather than simple point predictions. 
The conformal predictor is \emph{valid} in the sense that the accuracy of its predictions is guaranteed to meet the confidence level,  only assuming exchangeability in the data.
Since accuracy is guaranteed, the performance of a conformal predictor is measured through the \emph{efficiency} of the prediction sets.
Typically, a conformal predictor is built on an underlying machine learning algorithm and hence its predictive power is inherited from this algorithm. 
However, since the underlying machine learning algorithm is not trained with the objective of minimizing predictive efficiency it means that the resulting conformal predictor may be sub-optimal and not aligned sufficiently to this objective.
Hence, in this study we consider an approach to train the conformal predictor directly with maximum predictive efficiency as the optimization objective, and we focus specifically on the inductive conformal predictor for classification.
To do this, the conformal predictor is approximated by a differentiable objective function and gradient descent used to optimize it. 
The resulting parameter estimates are then passed to a proper inductive conformal predictor to give valid prediction sets.
We test the method on several real world data sets and find that the method is promising and in most cases gives improved predictive efficiency against a baseline conformal predictor.
\end{abstract}
 
\section{Introduction}

Conformal predictors (CP) are a form of machine learning algorithm that produce prediction sets, unlike most machine learning algorithms that provide point predictions. A prediction is correct if the true label is a member of the prediction set. Because prediction sets can be of variable sizes and in the extreme case can contain all possible labels, it is possible to construct an algorithm to control the accuracy of predictions. Indeed, under the simple assumption of exchangeability, for which, independently and identically distributed (i.i.d.) is a special case, it has been shown that CP is \emph{valid} in the sense that accuracy is guaranteed to follow a user-defined confidence level \citep{Vovk2005AlgorithmicWorld}.
This property is important since the conformal predictor is able to hedge its prediction, in order to achieve reliability. There are several decision support problems where this would be valuable. For example, if a medical doctor is faced with diagnosing the disease for a specific patient, a prediction that it could be one of two possible diseases with 95\% confidence made by CP may be more useful than a machine learning prediction of just one disease for which the reliability of the prediction is uncertain. The CP narrows down the options that the doctor can use to explore the correct diagnosis further \citep{Nouretdinov2014}.

Classification by CPs (conformal classiciation) is the subject of this article. There have been many examples where conformal classification has been applied successfully in real-world problems.
For example, \cite{NOURETDINOV2011809} use a transcendental CP for labelling processed MRI image processing using TCP to discover markers for depression diagnosis and prognosis. With a small sample size and at confidence level 90\% they are able to predict 22 out of 38 with singleton prediction sets (i.e. predicting a single label). 
\cite{Norinder2016} use an aggregated CP for binary classification of a large data set of estrogen receptor data, based on an underlying support vector machine. They found that using CP improved overall performance compared to traditional methods.
\cite{Wang2019} use aggregated CP to provide reliable predictions for 10 different species of a medical herb, dendrobium, using data sourced from an electronic nose.
\cite{Kagita2017} use conformal classification to construct a recommender system with confidence, by using a novel nonconformity measure to take account  precedence probability of each potential recommendation.
These recent examples show there is practical benefit in exploring the optimization of performance of conformal classifiers.

Since accuracy is controlled by the confidence level, the performance of the CP can be measured by the \emph{efficiency} of the predictions, which is, in general, some measure of the information provided by the prediction. There are several alternative measures of efficiency or inefficiency \citep{Vovk2016}.
For this study, the \emph{N criterion} \cite{Vovk2016}, which is the size of the prediction set, as the inefficiency measure, since this measure is directly related to the output of the CP when it is used in practice and is observed by the end-user of the CP. Very broadly, a smaller prediction set is more informative than a larger prediction set. The following example illustrates the informativeness of different prediction sets.
\begin{example} \label{ex:ineff}
Suppose there is a multiclass problem to classify people into one of four groups: $A, B, C$ and $D$.
\begin{itemize}
\item If a prediction set for a new person is output with a single response, e.g. $\{B\}$, then this is the most informative response.
\item The prediction set $\{B,D\}$ is less informative since it predicts the label will be either $B$ or $D$. However, it provides some information since it excludes the other two possibilities.
\item On the other hand, the prediction set $\{A,B,C,D\}$ provides no new information since it simply states that the person is one of the four groups we already know about. However, trivially $\{A,B,C,D\}$  will always be a correct prediction.
\end{itemize}
This example shows that the most to least informative prediction sets are ordered by increasing prediction set size.
\end{example}
Other inefficiency measures can be proposed as functions of prediction set size for different application objectives and an  alternative, log of prediction set size, is considered in this study to demonstrate our method works with different performance measures.

CPs are contrasted with traditional machine learning algorithms in Table \ref{tab:cpcontrast}.
\begin{table}[ht]
    \centering
    \begin{tabular}{|l|l|l|}
    \hline
         &  Traditional & \\
         &  machine learning & Conformal predictor (CP) \\
         \hline
  Type of prediction & Point prediction & Prediction set \\
  \hline
  Accuracy & Maximized & Fixed \\
  & & (confidence level) \\
  \hline
  Predictive efficiency & Fixed & Maximized \\
  (e.g. size of prediction) & & (e.g. minimize prediction set size)\\
  \hline
    \end{tabular}
\caption{Contrast conformal predictor with traditional machine learning.} 
\label{tab:cpcontrast}
\end{table}

Typically CPs have been built based on an underlying predictive algorithm P, hence their performance is reliant on how well P works. However, the objective function of P is not aligned with the CP objective of maximizing predictive efficiency. Therefore a CP built in this way is likely to be sub-optimal and misaligned to the goal of maximizing efficiency. This motivates research to build a CP by directly maximizing predictive efficiency. A direct approach to training CP for a probabilistic measure of efficiency is given by \cite{colombo2020}. However, a direct approach to training CP is not straightforward because the expression of CP as an optimization problem involves multiple step functions. In this article we consider an alternative strategy whereby the CP objective is approximated by a differentiable objective function to which gradient descent can be applied. This process is call \emph{surrogate conformal predictor optimization} (SCPO). 
The use of gradient descent to optimize prediction intervals has been applied successfully by \cite{Pearce2018} for regression problems outside the context of CP.
Because SCPO is an approximation to CP, it means it is not inherently valid. However, the parameters estimated by SCPO are passed to a CP which is valid. Again, because SCPO is an approximation, it will still be sub-optimal. However, since SCPO mirrors CP closely, we hypothesize that it can provide more efficient solutions than the classical use of CP with an underlying machine learning algorithm. We test this claim with experiments across several real world data sets.

The use of the N criterion as inefficiency measure means that the measure is dependent on the confidence level that is set, since as confidence level increases, we have a general increase in prediction set size to accommodate the higher confidence level.
This does mean that a separate optimization problem needs to be set for each confidence level we choose. This raises the question whether different confidence levels lead to different optimally trained CPs. This question is also explored in the experimental section.

We specifically focus on inductive conformal predictors (ICP) since they are the most popular form of CP for batch machine learning. 
Additionally, we consider only the classification problem where the outcome variable is a factor with finite possible labels, as opposed to regression.
The problem of implementing SCPO with ICP for regression has been studied in the linear case by \cite{Bellotti2020}.
However, the formulation of conformal classification is quite different to conformal regression, with a distinct algorithm for constructing the prediction sets. Hence further work has been conducted to implement this approach for classification and demonstrate that it works in this case.

The scope of this article is to introduce the methodology in Section \ref{methodology} and demonstrate its effectiveness with a linear model structure across several classification problems in Section \ref{experiments}. It is shown that in the majority of cases, SCPO improves performance of ICP. Also, we find that ICPs optimized for different confidence levels or inefficiency measures give different results. 
Conclusions are given in Section \ref{conclusion}.

\section{Methodology} \label{methodology}
We set up the ICP framework following \cite{Vovk2005AlgorithmicWorld}.
\begin{itemize}
\item Let $\varepsilon \in (0,1)$ be a user-defined significance level, and $1-\varepsilon$ is the confidence level for predictions.
\item Let $\bs{z}_1, \cdots, \bs{z}_n$ be a sequence of examples $\bs{z}_i = (\bs{x}_i, y_i)$ with vector of $m$ predictor variables $\bs{x}_i \in \mathbb{R}^m$ and response variable $y_i \in \{1,\cdots,C\}$.
\item Without loss of generality, let $1$ to $k$ index training data, $k+1$ to $l$ index calibration data and $l+1$ to $n$ index test data, for $1<k<l<n$.
\item A conformity measure (CM) is any computable function 
$\mathcal{A}(\bs{z}_1, \cdots, \bs{z}_k,  (\bs{x}, y) )$
such that $ \mathcal{A}$ is exchangeable: 
i.e. for all permutations $\pi$ of $1, \cdots, k$,
$$ \mathcal{A}( \bs{z}_1, \cdots,  \bs{z}_k,  (\bs{x}, y) ) 
=  \mathcal{A}(\bs{z}_{\pi(1)}, \cdots,  \bs{z}_{\pi(k)},   (\bs{x}, y) ). $$ 
\end{itemize}
In this study, we will consider optimizing the CM for minimizing inefficiency, therefore we use a parametric CM for some vector of $p$ parameters $\theta \in \mathbb{R}^p$,
$$\ap \textrm{ where } \theta = M(\bs{z}_1, \cdots,  \bs{z}_k)$$
 such that $M$ is an exchangeable function: i.e. for all permutations $\pi$ of $1, \cdots, k$,
$M( \bs{z}_1, \cdots,  \bs{z}_k ) =  M(\bs{z}_{\pi(1)}, \cdots,  \bs{z}_{\pi(k)})$.
The function $M$ is intended to be a model structure within which $\theta$ can be computed based on data $\bs{z}_1, \cdots,  \bs{z}_k$.
This is a typical form of CM in the literature where classically $M$ is some underlying parametric classification algorithm such as linear support vector machine, logistic regression or artificial neural network. So, e.g, using logistic regression as the underlying algorithm, $\theta$ corresponds to the model coefficients, whereas usng neural network, $\theta$ corresponds to the neural network weights. In this study, we consider parametric forms of CM but an underlying classification algorithm for computing them is unnecessary. Instead, a loss function associated with predictive inefficiency is directly minimized with respect to $\theta$.
The exchangeability condition on $M$ is uncontrovertial and most classification algorithms are invariant to order of training data.
\begin{itemize}
\item Finally, let $\ai = \apii$ denote CM for observation $i$.
\end{itemize}

The inductive conformal predictor (ICP) gives the prediction set at confidence level $1-\varepsilon$,
\begin{equation} \label{eq:predinterval1}
\Gamma^{\varepsilon}(\bs{x}) 
	= \left\{y \in  \{1,\cdots,C\} :  \sum_{j=k+1}^l {\mathbb{I} \left[\ap \ge \aj  \right]} + 1 
  >  \varepsilon (l-k+1) \right\}
\end{equation}
where $\mathbb{I}$ is the indicator function. \\
Assuming only that calibration and test data, $\bs{z}_{k+1}, \cdots, \bs{z}_n$, are exchangeable, ICP predictions are valid; i.e. for all $i \in \{l+1,\cdots,n\}$, 
\begin{equation} \label{eq:validity}
\mathbb{P}(y_i \in \Gamma^{\varepsilon}(\bs{x}_i) ) \ge 1-\varepsilon.
\end{equation}
Notice that (\ref{eq:predinterval1}) can be rewritten as
\begin{equation} \label{eq:predinterval2}
\Gamma^{\varepsilon}(\bs{x}) = \left\{y \in  \{1,\cdots,C\} :  \ap> q \right\}
\end{equation}
where $q$ is the $\varepsilon$ quantile of the calibration data set, $\alpha_{k+1}(\theta)$ to $\alpha_l (\theta)$ \footnote{
More precisely, if the CMs on the calibration data set are ordered, $\alpha_{(1)}(\theta) \le \alpha_{(2)}(\theta) < \cdots < \alpha_{(l-k)}(\theta)$, $q=\alpha_{(j)}(\theta)$ where $j=\lfloor \varepsilon(l-k+1) \rfloor - 1$.
}.

The inefficiency of a prediction set is measured as some function of the size of the prediction set,
$f(|\Gamma^{\varepsilon}|)$. The identity function is an obvious choice for $f$, the N criterion \citep{Vovk2016}, but others are possible as discussed in Section \ref{pm} below.
The optimization problem is then to minimize mean inefficiency across the training data,
$$\frac{1}{k} \sum_{i=1}^k {f \left( \sum_{y=1}^C {\mathbb{I} \left[\api > q \right]} \right)}$$
whilst holding accuracy approximately at the given confidence level to express validity. Note that the inequality in (\ref{eq:validity}) accounts for ties in the CM. If we choose a CM with sufficent granularity the number of ties will be minimal and the accuracy will be approximately equal to confidence level. Alternatively we can deploy a smoothed CP to ensure exact validity (see \cite{Vovk2005AlgorithmicWorld}). The validity constraint, as manifest in the training data, is written as
\begin{equation} \label{eq:valcon}
\frac{1}{k} \sum_{i=1}^k {\mathbb{I} \left[ \ai > q \right]} \approx 1- \varepsilon
\end{equation}
since $y_i \in \Gamma^{\varepsilon}(\bs{x}_i) \Leftrightarrow \mathbb{I}\left[ \api > q \right]$ from (\ref{eq:predinterval2}).
This approximation can be expressed in the loss function through a simple square loss term, following \cite{Pearce2018}. 
This is a departure from proper ICP since validity is an inherent property of ICP. However, this is not the case with SCPO where our intention is to impose \emph{empirical validity} through the loss function. In this sense, SCPO mimics ICP.

This process leads to the full loss function,
$$L_S(\theta) = \frac{1}{k} \sum_{i=1}^k {f \left( \sum_{y=1}^C {\mathbb{I} \left[ \api > q \right]} \right)}  
+ \lambda  {\left( \frac{1}{k} \sum_{i=1}^k {\mathbb{I} \left[ \ai > q \right]} - (1- \varepsilon) \right)}^2$$
where $\lambda>0$ controls the relative importance of the two objective components. 
The loss $L_S(\theta)$ is minimized with respect to $\theta$.
To avoid the indicator functions, we use approximation by the expit sigmoid function, 
$\mathbb{I}[x>0] \approx  \expit{x}$ for sufficiently large $\gamma$.
Larger values of $\lambda$ and $\gamma$, the closer SCPO mimics ICP. However, larger values for these hyperparameters may yield objective functions that are not so easy to optimize using gradient descent. This trade-off is explored in the experimental results through grid search.

The quantile term $q$ remains in the loss function $L_S$. In the ICP this would be computed from the calibration set. For the SCPO approximation, we have the option to estimate $q$ as part of the optimization problem. Constraint (\ref{eq:valcon}) ensures it emerges as approximately the $\varepsilon$-quantile.
However, we observe that CP is \emph{invariant} to the scale of the CM, so if they are all rescaled by the same factor, CP would behave the same. Hence, so long as the CM contains a parameter that is able to change the scale, $q$ can be set to a constant value and the CM will rescale to meet this target through constraint (\ref{eq:valcon}).
This then leads to the loss function approximation,
$$L(\theta) = \frac{1}{k} \sum_{i=1}^k {f(s_i)}  + \lambda  V^2$$
where
$$s_i = \sum_{y=1}^C {\sigma{(\gamma (\api - q))}},$$
$$V = \frac{1}{k} \sum_{i=1}^k {\sigma{(\gamma(\ai - q))}} - (1- \varepsilon).$$
To minimize the loss function, gradient descent is used with gradients,
$$\pd{L(\theta)}{\theta_j} = \frac{1}{k} \sum_{i=1}^k {f'(s_i) \pd{s_i}{\theta_j}}  + 2 \lambda  V \pd{V}{\theta_j}$$
where
$$\pd{s_i}{\theta_j} = \gamma \sum_{y=1}^C 
{\sigma{(\gamma (\api - q))}  \left( 1- \sigma{(\gamma (\api - q))} \right) 
 \pd{\api}{\theta_j} },$$
$$\pd{V}{\theta_j} = \frac{\gamma}{k} \sum_{i=1}^k 
{\sigma{(\gamma(\ai - q))} \left( 1-\sigma{(\gamma(\ai - q))} \right) \pd{\ai}{\theta_j}}$$
and $\sigma(a)={(1+\exp a)}^{-1}$.
This now just depends on providing a differentiable CM.

In practice, the loss function $L(\theta)$ typically has too many extreme values amongst flat plateaus and is therefore not easy to traverse in gradient descent. Therefore a monotonically increasing transformation can be used to reduce the extreme peaks whilst having the same mathematical solution. An obvious choice is $\log L(\theta)$ function which gives gradients,
$$\pd{\log L(\theta)}{\theta_j} = \pd{L(\theta)}{\theta_j} \cdot \frac{1}{L(\theta)} .$$
However, our study shows that even this log transform does not give the best results, in general, and the transformation $-L(\theta)^{-1}$ is also considered, giving the gradients,
$$\pd{\left( -L(\theta)^{-1}\right)}{\theta_j} = \pd{L(\theta)}{\theta_j} \cdot \frac{1}{L(\theta)^2} .$$

For ease of computation, the expressions in this section are converted to matrix notation shown in the Appendix. For this study they were implemented in R.

\subsection{Multiclass Linear Conformity Measure}

For this study we use a CM with linear model structure. This is perhaps the simplest model structure we can choose and allows the methodology to be demonstrated without further complications linked to the CM, as a proof of concept. However, for the experiments introduced in Section \ref{experiments}, it is not suggested that a linear model is the best model structure. 

\begin{figure}[ht]
\centering
\includegraphics[scale=0.6] {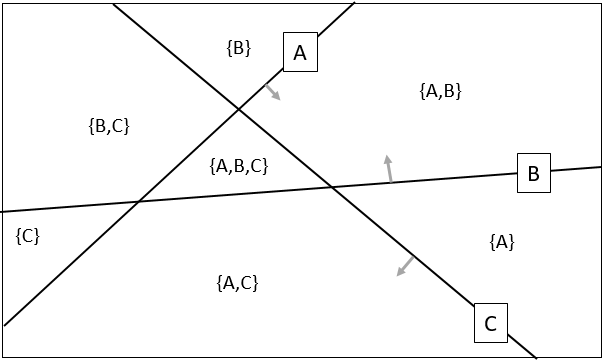}
 \caption{Illustration of the multiclass linear conformity measure with 2 variables (horizontal and vertical axes). The lines labelled A, B and C show $ \theta_{[y]} \cdot \bs{x} = q$ for each class label $y \in \{A,B,C\}$, with the small arrow showing the direction for  $\theta_{[y]} \cdot \bs{x} > q$. Consequently, each area between the lines corresponds to a different prediction set for any example falling inside that area.} 
\label{fig:mlcm2d}
\end{figure}

Take the case of a multiclass linear parameterization $\theta = (\theta_{[1]}, \cdots, \theta_{[C]})$ where each $\theta_{[y]}$ is a sub-vector of parameters for each class label $y$ and
\begin{equation} \label{eq:linear_params}
A(\bs{x}, y) = \theta_{[y]} \cdot \bs{x}
\end{equation}
and an intercept is introduced by including a column of 1's in $\bs{x}$.
Then
$$\pd{A(\bs{x}, y)}{\theta_j} = \left\{ 
\begin{array}{ll}
x_a & \textrm{ if } y=y' \\
0 & \textrm{otherwise}
\end{array}
\right.$$
where parameter $a$ of $\theta_{[y']}$ corresponds to $\theta_j$.
Figure \ref{fig:mlcm2d} illlustrates the multiclass linear CM with just 2 variables and a 3 class label problem where each partition expresses a different prediction set. 

Since this multiclass linear CM can be given at different scalings, and express as the same ICP, by multiplying all $\theta_j$ by the same constant factor, $q$ can be set to any non-zero real number, as discussed earlier. In this study, for simplicity, we use $q=1$.

\subsection{Performance measures} \label{pm}

Performance of the SCPO-ICP method is measured on an independent test set by accuracy,
$$\textrm{Acc.} = \frac{1}{n-l} \sum_{i=l+1}^n {\mathbb{I}(y_i \in \Gamma^\varepsilon(\mathbf{x}_i))}$$
which should be approximately $1-\varepsilon$ if the algorithm is valid, and mean inefficiency,
$$\textrm{Ineff.} = \frac{1}{n-l} \sum_{i=l+1}^n {f(|\Gamma^\varepsilon(\mathbf{x}_i)|)}$$
which we hope to minimize.
The choice of $f$ depends largely on the application. If the cost of a prediction set increases with its size, then the identity function $f(w)=w$ is natural. However, for some applications, we may suppose that the cost of an increase in prediction set size reduces with its size. For example, the cost difference from a prediction of 1 single label to 2 may be high, whereas the utility gain of a prediction of 5 labels against 6 may not be very high, relatively. This would suggest the log function $f(w)=\log(w+1)$.
In this study, we consider both options for contrast. There are many other options, e.g. a square function, that could be considered for other applications.

Performance of SCPO-ICP is measured against a traditional ICP using a CM based on a classification algorithm. Since in this study, we use a linear CM, we therefore take a linear multi-class classification algorithm as the baseline algorithm. Multinomial logistic regression is the classical statistical model for multi-class classification so we take it as our choice for baseline. The trained multinomial regression algorithms will output the probability of each label conditional on a new example,  hence this is taken as the value of the CM, $A(\mathbf{x},y)=\mathbb{P}(Y_i=y | \mathbf{x})$ for multinomial regression. The multinomial function \texttt{multinom} in the \texttt{nnet} library is used in R for the baseline ICP.

\section{Experiments} \label{experiments}

Experiments using the SCPO method with ICP are conducted on several data sets. 

\subsection{Data sets}

Seven data sets expressing classification problems are sourced from the UCI Machine Learning data repository \citep{Frank2010UCIRepository}, except the ``dna'' and ``mice'' data sets:
the ``mice'' data set which  is downloadable as the \emph{mice protein classification} data set on www.kaggle.com and the``dna'' data was provided on Statlog by Ross King at Strathclyde University \footnote{Now available in R: see www.rdocumentation.org/packages/mlbench/versions/2.1-1/topics/DNA}
. All data sets used in this study are listed in Table \ref{tab:datasets}. Only data sets with more than 1,000 observations were used so that a good number of observations are available for training, calibration and test sets. Data sets express multi-classification problems, except for ``wine' and ``polish''' which are set up as a binary classification.  
Separate notes for each data set are given below.
\begin{itemize}
\item The downloaded ``avila'' data set has 12 labels, but 4 have relatively low sample size ($<300$) and so they have been taken together as a single \emph{Others} label.
\item The ``covtype'' data set has been reduced in size so that computation time is manageable for the intensive grid search.
\item The outcome variable given with the ``wine'' data set is an integer between 0 and 10 indicating wine quality. However, we deliberately want to use it as a balanced binary classification problem, hence we use the outcome variable given by wine quality $\ge 6$ to mean ``good quality''. This gives 63\% good quality wines versus 37\% poor quality.

\item The outcome variable for the downloaded ``polish'' data set is imbalanced since bankruptcy is a rare event. Hence to ensure balanced classes during experiments the non-bankruptcy firms are undersampled. This leaves 66\% non-bankruptcies versus 34\% bankruptcies.

\item Only the protein expression data in ``mice'' are used for prediction. There are many missing values in this data set. Hence 3 observations with completely missing data are deleted, 7 variables with more than 100 missing values are deleted, and mean imputation is used for the remainder.
\end{itemize}

Allocation of examples to training, calibration and test set is done by random sampling, except for ``avila'' which is delivered as training and test, so just the training data is randomly split into proper training and calibration data sets and ``fotp'' where the data is already divided into the three parts.

All data sets are normalized prior to using gradient descent since gradient descent can work poorly on unnormalized data. All variables are normalized to mean 0 and variance 1. To avoid bias in performance of the test data, all norms for the rescaling are computed only from the training data. Note that the downloaded data sets ``fotp'' and ``dna'' are already normalized.

Class imbalance is a consideration for these experiments since a CP can perform well simply by predicting the majority class. For example, if one label accounts for 80\% of observations, then singleton prediction sets can be constructed at 80\% confidence level simply by predicting this label. However, as can be seen in Table \ref{tab:datasets}, none of the data sets as used in these experiments have extreme imbalance in the majority class label.
\begin{table}[ht]
\centering
\begin{tabular}{l p{3.7cm} l l l l l l}
  \hline
Data set & Description & \#var &  \#labels & \#Train & \#Cal & \#Test & \%maj. \\ 
  \hline
avila & Identify Avila Bible text copyist. & 10 & 9 & 6954 & 3476 & 10437 & 41 \\
\hline
covtype & Identify tree cover type for forest location. & 54 & 7 & 12000& 12000& 12000 & 48 \\
   \hline
dna & Identify nucleotides for a sample of DNA splice junctions. & 180 & 3 & 667 & 666 &667 & 52 \\
\hline
fotp & Predict the best heuristic for a first-order theorem prover. & 51 & 6 & 3059&1530 &1529 & 42 \\
   \hline
wine &Predict wine quality. & 64 & 2 & 2297 & 2100&2100 & 63 \\
   \hline
polish & Polish bankruptcy data. & 60 & 2 & 444 & 444 & 444 & 66 \\
\hline
mice & Classification from mice protein expression. & 71 & 8 & 359 & 359 & 359 & 15 \\
\hline
\end{tabular}
\caption{Data sets sourced from UCI Machine Learning data repository. 
\#var = number of predictor variables, 
\#labels = number of unique class labels,
\#Train, \#Cal, \#Test are number of observations in the training, calibration and test sets, respectively,
\% maj. = percentage of labels in the majority class.} 
\label{tab:datasets}
\end{table}

\subsection{Experimental set up}
For each experiment, the optimal values of $\lambda$ and $\gamma$ are chosen by grid search across values of $\lambda \in \{10, 100, 500, 5000, 10000 \}$, $\gamma \in \{1, 2, 5, 10\}$ and gradient descent learning rate $\eta \in \{1, 10, 100,1000 \}$ for each significance level $\varepsilon \in \{0.2, 0.1, 0.05,$ $0.025, 0.01\}$. 
The set of hyperparameters that minimized the loss function $L(\theta)$ on training data were used for the ICP. Since the ICP used to compute loss within the grid search only has access to the traing data, the training data is also used for both calibration and test data. Clearly this will lead to overfitting and violates the exchangeability assumption, so the internal grid search ICP is not valid in the sense of (\ref{eq:validity}). However this is not a problem since, in this experiment, it does not need to be valid: it passes hyperparameters to a further stage of ICP that uses an independent calibration and test set and hence is valid.
We find that the range of hyperparameter values given above were generally successful for minimizing the loss function. 
We expect that a search across a wider range of values would show some small improvement in performance but the grid search already involves 400 iterations for each data set which already takes a long time to compute and so for practical reasons it was not expanded.

The set of hyperparameters that minimize inefficiency when running a \emph{training ICP} with these hyperparameters on just training data, 
i.e. training data is also used for both calibration and testing. This is done so that when the final ICP is used based on the output of SCPO and grid search, there is no overfitting and the ICP is valid. 
Understandably the training ICP is not valid since the use of the same (training) data set in all three roles violates the exchangeability condition. However, we find that the outcome of the training ICP was a good indicator for the performance on the true test data, which is what matters for hyperparameter selection. Validity only matters for the final ICP.

Larger values of $\varepsilon$ are not considered since, firstly, for most practical purposes it is larger confidence levels ($1-\varepsilon$) that will be useful for the end-user
and, secondly, the ICP can often meet low confidence levels simply by predicting the most common class label (e.g. suppose the most common class label occurs 25\% of the time, then to meet a confidence level of 25\% or lower, just predict that class label) so this is not an interesting case.

For all experiments, the initial parameter values for gradient descent are set to zero.

\subsection{Results}

As discussed in Section \ref{methodology}, several transformations of the loss function are explored to determine the best performance with gradient descent. We find that typically  $L(\theta)$ does not give good performance due to extremes and plateaus in the objective function. Across the seven data sets, based on a separate grid search for each Loss transform so that the best hyperparameters are chosen specifically for each case, $L(\theta)$ never gives best performance, $\log L(\theta)$ gives good performance for 2 data sets, $-L(\theta)^{-1}$ in 4 cases and $-L(\theta)^{-2}$ in just 1 case.  Results for two example data sets are shown in Table \ref{tab:Lpower}.
For this reason we use $-L(\theta)^{-1}$ as transformation of the Loss function.
\begin{table}[ht]
\centering
\begin{tabular}{l lll lll}
\hline
Data set & \multicolumn{3}{c}{``avila''} & \multicolumn{3}{c}{``fotp''} \\
  \hline
Loss&Training & \multicolumn{2}{c}{Test} &Training & \multicolumn{2}{c}{Test} \\
transform & Ineff. & Acc. (\%) & Ineff. & Ineff. & Acc. (\%) & Ineff. \\ 
  \hline
$L(\theta)$ & 2.938 & 94.8 & 3.102 & 4.641 & 94.6 & 4.933 \\
$\log L(\theta)$ & 2.874 & 94.8 & 3.062 & 3.932 & 94.9 & 4.614 \\
$-L(\theta)^{-1}$ &2.852 & 94.8 & 3.042 & 3.981 & 95.0 & 4.529 \\
$-L(\theta)^{-2}$ &2.823 & 94.7 & 3.038 & 3.979 & 95.6 & 4.927 \\
   \hline
\end{tabular}
\caption{Effect of different transformations of the loss function, with $\varepsilon=0.05$ and inefficiency measure $f(w)=w$.}
\label{tab:Lpower}
\end{table}

Main results across all data sets are shown in Tables \ref{optcp:mainresults_all} and \ref{optcp:mainresults_logw}. They show results with inefficiency measured by mean size of the prediction set, i.e. $f(w)=w$, and by log of mean size, $f(w)=\log(w+1)$, respectively. 
The tables show that test accuracy approximates the confidence level closely in all cases, demonstrating that the SCPO-ICP method is valid.

Both tables show that the SCPO approach gives improvement over the baseline ICP method with regard to Inefficiency, as shown by Ch (percentage change between baseline and SCPO inefficiency) $>0$ in the majority of cases.
 In general, worse performance is seen at the 99\% confidence level for several data sets. 
The exception to this is ``mice'' for which worse performance for SCPO is seen at lower confidence levels and is 4.1\% worse than baseline for $f(w)=w$ at 80\% confidence level. However, this is an unusual case since mean prediction set width is less than 1, which suggests that many prediction sets are empty, which is also not desirable. 
For ``fotp'' we see negative results at the 99\% confidence level although it is by a small amount ($<$1\%) and this is not found to be statistically significant using a binomial test
\footnote{The statistical test used here is to label test examples 0 if the baseline ICP gives smaller inefficiency and 1 if SCPO gives smaller inefficiency, discarding those cases where they are the same. If the two methods perform equally well then we expect that the number of 0's will be the same as 1's and this is tested using the binomial test.}
 (p-value=0.206 and 0.027 for the identity and log inefficiency measures, respectively).
The method performs less well for the ``covtype'' data set across all confidence levels, but only by up to 5.1\% worse than baseline at 99\% confidence levels. Using the binomial test, these results are statistically significant (p-value$<$0.001 for all cases).

These results demonstrate that the method is not universely successful. 
Since the model structure of the CM for SCPO is more general than for the baseline using multinomial regression, there must exist a CM solution in SCPO which at least matches the baseline, but this optimum is not being found. 
The problem is either due to SCPO not being a sufficient approximation to ICP and/or the gradient descent optimization not performing well. Experience working with this problem suggests the latter, since manual intervention with the gradient descent can give improved performance. This suggests further work to improve the implementation of gradient descent for SCPO.

A very large improvement in performance using SCPO-ICP is reported for both inefficiency measures for the two data sets ``avila'', ``wine'' and ``polish'', and for ``mice'' at high confidence levels.
Interestingly, higher settings of $\lambda$ are selected in the grid search for higher confidence levels, although there is no perceivable pattern with $\gamma$.
\begin{singlespace}
\begin{table}[ht]
\centering
\begin{tabular}{ll|rr|rr|rr|r}
  \hline
&Confidence&&&\multicolumn{2}{c|}{SCPO-ICP} & \multicolumn{2}{c|}{Baseline} &\\
Dataset & level & $\lambda$ & $\gamma$ & Acc. & Ineff. & Acc. &Ineff. & Ch. \\ 
  \hline
avila & 80.0 & 100 & 2 & 80.1 & 1.81 & 79.9 & 2.04 & 11.0 \\ 
  avila & 90.0 & 500 & 1 & 90.1 & 2.51 & 90.2 & 3.33 & 24.7 \\ 
  avila & 95.0 & 1000 & 2 & 94.7 & 3.04 & 95.3 & 4.84 & 37.2 \\ 
  avila & 97.5 & 5000 & 2 & 97.1 & 3.55 & 97.8 & 6.21 & 42.8 \\ 
  avila & 99.0 & 5000 & 1 & 99.0 & 5.22 & 99.1 & 7.84 & 33.5 \\ 
   \hline
covtype & 80.0 & 100 & 2 & 80.0 & 1.19 & 79.8 & 1.16 & -2.3 \\ 
  covtype & 90.0 & 100 & 1 & 89.8 & 1.52 & 89.5 & 1.49 & -2.1 \\ 
  covtype & 95.0 & 500 & 2 & 94.8 & 1.81 & 94.8 & 1.77 & -2.2 \\ 
  covtype & 97.5 & 1000 & 2 & 97.5 & 2.03 & 97.4 & 2.00 & -1.7 \\ 
  covtype & 99.0 & 5000 & 1 & 98.9 & 2.39 & 98.8 & 2.27 & -5.1 \\ 
   \hline
dna & 80.0 & 10 & 1 & 83.4 & 0.89 & 80.5 & 1.01 & 12.1 \\ 
  dna & 90.0 & 10 & 2 & 92.8 & 1.08 & 89.4 & 1.26 & 14.4 \\ 
  dna & 95.0 & 10 & 2 & 97.6 & 1.22 & 95.1 & 1.52 & 19.5 \\ 
  dna & 97.5 & 10 & 1 & 98.8 & 1.30 & 97.2 & 1.76 & 26.1 \\ 
  dna & 99.0 & 100 & 2 & 99.6 & 1.69 & 98.5 & 1.97 & 14.1 \\ 
   \hline
fotp & 80.0 & 500 & 2 & 77.9 & 2.71 & 79.6 & 2.85 & 5.0 \\ 
  fotp & 90.0 & 500 & 1 & 89.7 & 3.87 & 90.9 & 4.07 & 4.9 \\ 
  fotp & 95.0 & 1000 & 1 & 95.0 & 4.53 & 96.5 & 4.94 & 8.3 \\ 
  fotp & 97.5 & 1000 & 1 & 97.6 & 5.18 & 98.3 & 5.52 & 6.2 \\ 
  fotp & 99.0 & 5000 & 1 & 99.3 & 5.75 & 98.9 & 5.72 & -0.6 \\ 
   \hline
wine & 80.0 & 100 & 2 & 78.4 & 1.10 & 80.8 & 1.79 & 38.8 \\ 
  wine & 90.0 & 100 & 2 & 89.3 & 1.40 & 91.6 & 1.91 & 26.6 \\ 
  wine & 95.0 & 500 & 2 & 94.5 & 1.57 & 95.9 & 1.96 & 20.0 \\ 
  wine & 97.5 & 1000 & 1 & 97.2 & 1.72 & 98.0 & 1.98 & 13.2 \\ 
  wine & 99.0 & 1000 & 2 & 98.7 & 1.83 & 99.3 & 1.99 & 8.3 \\ 
   \hline
polish & 80.0 & 100 & 5 & 81.5 & 1.20 & 79.3 & 1.74 & 31.0 \\ 
  polish & 90.0 & 100 & 1 & 87.6 & 1.40 & 89.4 & 1.87 & 25.2 \\ 
  polish & 95.0 & 500 & 2 & 92.3 & 1.57 & 95.0 & 1.94 & 19.3 \\ 
  polish & 97.5 & 500 & 2 & 95.7 & 1.74 & 100.0 & 2.00 & 13.0 \\ 
  polish & 99.0 & 5000 & 1 & 98.6 & 1.91 & 100.0 & 2.00 & 4.5 \\ 
   \hline
mice & 80.0 & 100 & 2 & 81.9 & 0.85 & 79.7 & 0.81 & -4.1 \\ 
  mice & 90.0 & 100 & 1 & 91.6 & 0.95 & 91.9 & 0.97 & 1.7 \\ 
  mice & 95.0 & 100 & 1 & 95.3 & 1.01 & 96.9 & 1.11 & 9.1 \\ 
  mice & 97.5 & 100 & 1 & 99.2 & 1.09 & 99.2 & 1.48 & 26.0 \\ 
  mice & 99.0 & 100 & 1 & 99.7 & 1.35 & 99.7 & 2.33 & 41.8 \\ 
   \hline
\end{tabular}
\caption{Accuracy and inefficiency on test data for each data set at different confidence levels, with inefficiency measure $f(w)=w$.
Change in inefficiency is Ch. = 1 - SCPO Ineff. / Baseline Ineff., as a percentage.} 
\label{optcp:mainresults_all}
\end{table}
\end{singlespace}
\begin{singlespace}
\begin{table}[ht]
\centering
\begin{tabular}{ll|rr|rr|rr|r}
  \hline
&Confidence&&&\multicolumn{2}{c|}{SCPO-ICP} & \multicolumn{2}{c|}{Baseline} &\\
Dataset & level & $\lambda$ & $\gamma$ & Acc. & Ineff. & Acc. &Ineff. & Ch. \\ 
  \hline
avila & 80.0 & 100 & 2 & 79.7 & 0.982 & 79.9 & 1.087 & 9.7 \\ 
  avila & 90.0 & 100 & 2 & 90.4 & 1.190 & 90.2 & 1.418 & 16.1 \\ 
  avila & 95.0 & 500 & 2 & 94.8 & 1.363 & 95.3 & 1.713 & 20.5 \\ 
  avila & 97.5 & 500 & 1 & 97.6 & 1.463 & 97.8 & 1.938 & 24.5 \\ 
  avila & 99.0 & 1000 & 2 & 99.0 & 1.772 & 99.1 & 2.163 & 18.1 \\ 
   \hline
covtype & 80.0 & 100 & 1 & 79.7 & 0.776 & 79.8 & 0.756 & -2.7 \\ 
  covtype & 90.0 & 100 & 2 & 90.1 & 0.917 & 89.5 & 0.892 & -2.8 \\ 
  covtype & 95.0 & 100 & 1 & 94.9 & 1.027 & 94.8 & 1.002 & -2.6 \\ 
  covtype & 97.5 & 500 & 2 & 97.4 & 1.105 & 97.4 & 1.086 & -1.7 \\ 
  covtype & 99.0 & 1000 & 2 & 99.1 & 1.222 & 98.8 & 1.171 & -4.4 \\ 
   \hline
dna & 80.0 & 10 & 1 & 82.9 & 0.614 & 80.5 & 0.696 & 11.8 \\ 
  dna & 90.0 & 10 & 1 & 92.7 & 0.701 & 89.4 & 0.795 & 11.8 \\ 
  dna & 95.0 & 10 & 1 & 96.7 & 0.764 & 95.1 & 0.893 & 14.5 \\ 
  dna & 97.5 & 100 & 2 & 98.4 & 0.825 & 97.2 & 0.979 & 15.7 \\ 
  dna & 99.0 & 100 & 2 & 99.4 & 0.935 & 98.5 & 1.054 & 11.3 \\ 
   \hline
fotp & 80.0 & 100 & 2 & 79.6 & 1.289 & 79.6 & 1.295 & 0.5 \\ 
  fotp & 90.0 & 100 & 1 & 89.9 & 1.511 & 90.9 & 1.582 & 4.5 \\ 
  fotp & 95.0 & 100 & 1 & 95.6 & 1.730 & 96.5 & 1.757 & 1.5 \\ 
  fotp & 97.5 & 500 & 1 & 97.7 & 1.833 & 98.3 & 1.862 & 1.5 \\ 
  fotp & 99.0 & 500 & 1 & 99.3 & 1.902 & 98.9 & 1.897 & -0.3 \\ 
   \hline
wine & 80.0 & 100 & 2 & 79.3 & 0.738 & 80.8 & 1.014 & 27.3 \\ 
  wine & 90.0 & 100 & 2 & 90.1 & 0.856 & 91.6 & 1.063 & 19.5 \\ 
  wine & 95.0 & 100 & 2 & 93.4 & 0.910 & 95.9 & 1.081 & 15.8 \\ 
  wine & 97.5 & 500 & 2 & 97.0 & 0.979 & 98.0 & 1.090 & 10.1 \\ 
  wine & 99.0 & 500 & 2 & 98.7 & 1.030 & 99.3 & 1.096 & 6.0 \\ 
   \hline
polish & 80.0 & 100 & 2 & 79.3 & 0.752 & 79.3 & 0.995 & 24.4 \\ 
  polish & 90.0 & 100 & 5 & 87.6 & 0.863 & 89.4 & 1.045 & 17.4 \\ 
  polish & 95.0 & 100 & 1 & 92.1 & 0.918 & 95.0 & 1.074 & 14.5 \\ 
  polish & 97.5 & 100 & 2 & 95.5 & 0.990 & 100.0 & 1.099 & 9.9 \\ 
  polish & 99.0 & 500 & 1 & 98.6 & 1.066 & 100.0 & 1.099 & 3.0 \\ 
    \hline
mice & 80.0 & 500 & 1 & 76.0 & 0.557 & 79.7 & 0.564 & 1.2 \\ 
  mice & 90.0 & 100 & 1 & 88.6 & 0.627 & 91.9 & 0.670 & 6.4 \\ 
  mice & 95.0 & 100 & 1 & 95.8 & 0.707 & 96.9 & 0.733 & 3.6 \\ 
  mice & 97.5 & 100 & 1 & 97.5 & 0.714 & 99.2 & 0.864 & 17.3 \\ 
  mice & 99.0 & 100 & 1 & 99.7 & 0.811 & 99.7 & 1.118 & 27.5 \\ 
   \hline
\end{tabular}
\caption{Accuracy and inefficiency on test data for each data set at different confidence levels, with inefficiency measure $f(w)=\log(w+1)$. Change in inefficiency is Ch. = 1 - SCPO Ineff. / Baseline Ineff., as a percentage.} 
\label{optcp:mainresults_logw}
\end{table}
\end{singlespace}

For the grid search, we used the inefficiency measured using the ICP on only the training data to select hyperparameters. This is a good indicator of the performance by the ICP predicting on the true test set.


Since the objective is to minimize inefficiency as some function of prediction set width, this will be dependent on the confidence level. We test how sensitive the method is to the confidence level when estimating parameter estimates in the CM. To do this we train the CM at one confidence level, 80\%, and apply to predictions made at a different confidence level, 95\%. Results are shown in Table \ref{tab:ineff8095}. They show that predictions based on training the CM at the 80\% confidence level are typically worse than the inefficiency that can be achieved when training at the proper confidence level (the only exception is ``wine'' for which both CPs give the same performance). The consequence of this is that problems at different confidence levels will typically require a CM parameterization that is different for each confidence level.
\begin{table}[ht]
\centering
\begin{tabular}{lrr}
  \hline
Data set & Ineff. (80\%)& Ineff. (95\%) \\ 
  \hline
avila & 4.92 & 3.04 \\ 
  covtype & 5.85 & 1.81 \\ 
  dna & 1.26 & 1.22 \\ 
  fotp & 5.25 & 4.53 \\ 
  wine & 1.57 & 1.57 \\ 
 polish & 1.63 & 1.57 \\ 
  mice & 4.05 & 1.01 \\ 
  \hline
\end{tabular}
\caption{Results training at 80\% and 95\% confidence level (columns 2 and 3 respectively) and applying to test data at 95\% confidence level, with inefficiency measure $f(w)=w$.} 
\label{tab:ineff8095}
\end{table}



\section{Conclusion} \label{conclusion}

In this study, we have proposed a method to approximate the ICP for classification with a differentiable objective function. Gradient descent was then used to determine optimal parameters for a linear conformity measure. Preliminary experiments over 5 real world classification problems have demonstrated the effectiveness of this method to reduce the inefficiency of the prediction sets as measured by a function of the width of a prediction set, relative to a classical ICP based on an underlying multinomial logistic regression model. In one data set ``covtype'', we did not see an improvement in performance which we expect is a limitation with the current optimization algorithm which requires further study. 
We employed a simple grid search and a simple gradient descent controlled by a constant learning rate.
The results suggest further research to improve search over hyperparameters and the use of alternative gradient descent algorithms, such as allowing for random variation in the starting position and an adaptive learning rate.
Nevertheless, the proposed SCPO approach shows best results in 4 out of 5 of the data sets.

Further experiments demonstrate that each confidence level requires a distinct set of parameter estimates to perform well. In other words, in general, for any single problem, there is no single conformity measure that will work well at all confidence levels.

Since this paper introduces the SCPO methodology for conformal classification, we use a linear conformity measure as a proof of concept. However, further research work is required to extend it to more complex model structures for the conformity measure that can be estimated using gradient descent, such as artificial neural networks.

\section*{Acknowledgement}
Thank you to anonymous reviewers who commented on this article. I have made some revisions based on their suggestions in this preprint, but will make extended changes that are required for a future submission for publication.

\appendix
\section{Matrix notation}

Matrix notation for computation of the gradients is given here.
Operation is only on training data so observations are indexed $i \in {1, \cdots, k}$.
\begin{itemize}
\item Let $\bs{X}$ be $(k \times m)$ matrix of predictor variable values.
\item Let $\bs{Y}$ be column matrix of $k$ labels.
\item Let $\bs\theta$ be a column matrix of $p$ parameter corresponding to $\theta$.
\item Let $\bs{A}$ be the $(k \times C)$ matrix of $A(\bs{x}_i,y)$, 
with each column corresponding to values of $y \in \{1, \cdots, C\}$, 
computed from $\bs{X}$ and  $\bs\theta$.
\item Let $\bs{A}'^{[i]}$ be the $(C \times p)$ matrix of $\pd{A(\bs{x}_i, y)}{\theta_j}$  for each observation $i$,
computed from $\bs{A}$, $\bs{X}$ and  $\bs\theta$.
\item Let $\bs\alpha$ be the column matrix of $\alpha_i$. 
Then $\bs\alpha_i = \bs{A}_{iy}$ where $y=\bs{Y}_i$. 
\item Let $\bs\alpha'$ be the $(k \times p)$ matrix of $\pd{\alpha_i}{\theta_j}$.
Then $\bs\alpha'_{ij} = \mathbf{A}'^{[i]}_{yj}$ where $y=\bs{Y}_i$. 

\item Let $\bs\sigma = \sigma(\gamma(\bs{A}- q\bs1))$
and $\bs\sigma^{[\alpha]} = \sigma(\gamma(\bs\alpha- q\bs1))$.
\item Then $\bs{S} = \bs\sigma \bs1$ is the column matrix of $s_i$.
\item $V = \frac{1}{k}  \bs1 \bs\sigma^{[\alpha]} - (1-\varepsilon)$.
\item $L = \frac{1}{k} \bs1 f(\bs{S}) + \lambda V^2$ expresses  $L(\theta)$.

\item Let $\bs{S}'$ be the $(k \times p)$ matrix of $\pd{s_i}{\theta_j}$.
Then $\bs{S}'_{i \cdot} = \gamma {[\bs\sigma * (\bs1 - \bs\sigma) ]}_{i \cdot} \bs{A}'^{[i]}$.

\item Let $\bs{V}' = \frac{\gamma}{k} \bs\alpha'^T [\bs\sigma^{[\alpha]} * (\bs1 - \bs\sigma^{[\alpha]}) ]$ be the column matrix of $\pd{V}{\theta_j}$.

\item Let $\bs{L}' = \frac{1}{k} \bs{S}'^T f'(\bs{S}) + 2 \lambda V \bs{V}'$ be the column matrix of gradients $\pd{L(\theta)}{\theta_j}$.
\item Let  $\bs{L}_{\log}' =  \bs{L}'  / L$ be the column matrix of gradients $\pd{\log L(\theta)}{\theta_j}$.
\end{itemize}

For the specific multi-class linear NCM given in (\ref{eq:linear_params}), the number of parameters $p=Cm$ and
\begin{itemize}
\item Let $\bs{T}$ be a $(m \times C)$ matrix of parameters such that $\bs{T}_{jy} = \bs\theta_{j+m(y-1)}$, 
and so the $y$th column represents $\theta_{[y]}$.
Then $\bs{A} = \bs{X}\bs{T}$.
\item $\bs{A}'^{[i]}_{yl} = \left\{ \begin{array}{ll}
\bs{X}_{ij} & \textrm{ if } l = j + m(y-1) \textrm{ for some } j  \in \{1, \cdots, m\} \\
0 & \textrm{otherwise}
\end{array}
\right.$.
\end{itemize}
Notes: 
$\bs{M}_{i \cdot}$ refers to the $i$th row of $\bs M$;
$\bs1$ is a column matrix of 1s;
the operator $*$ is the element-by-element multiplication of two matrices.

\bibliographystyle{plainnat}
\bibliography{optcpc_v3}

\end{document}